\documentclass[a4paper,fleqn]{cas-dc}

\usepackage[authoryear]{natbib}
\usepackage{hyperref}
\usepackage{graphicx}
\usepackage{subfigure}
\usepackage{epstopdf}
\usepackage{color}
\usepackage[noend]{algpseudocode}
\usepackage{algorithmicx,algorithm}
\usepackage{caption}
\usepackage{tabularx}
\usepackage{xcolor}
\def\tsc#1{\csdef{#1}{\textsc{\lowercase{#1}}\xspace}}
\tsc{WGM}
\tsc{QE}
\tsc{EP}
\tsc{PMS}
\tsc{BEC}
\tsc{DE}
\hypersetup{
    colorlinks=true,            
    linkcolor=blue,              
    citecolor=blue,            
    filecolor=blue,   
    urlcolor=magenta            
}

\begin{document}
\captionsetup[figure]{labelfont={bf},name={Fig.},labelsep=period}
\captionsetup[table]{labelfont={bf}, labelsep=space }

\begin{sloppypar}
	\let\WriteBookmarks\relax
	\def\floatpagepagefraction{1}
	\def\textpagefraction{.001}
	\let\printorcid\relax
	\shorttitle{}
	\shortauthors{A.X. Ning et al.} 

	\title [mode = title]{Artistic-style text detector and a new Movie-Poster dataset}



	\author[1]{\textcolor[RGB]{0,0,1}{Aoxiang Ning}}
 	\address[1]{College Of Computer Science And Engineering,  Chongqing University of Technology, 400054, China}
	\ead{ningax@stu.cqut.edu.cn}
        \author[2]{\textcolor[RGB]{0,0,1}{Yiting Wei}}
        \ead{wyt18439532876@email.swu.edu.cn}
        \address[2]{College of Agronomy and Biotechnology, Southwest University, 400715, China}


	\author[1]{\textcolor[RGB]{0,0,1}{Minglong Xue}}

        \cormark[1]
	\ead{xueml@cqut.edu.cn}

        \author[3]{\textcolor[RGB]{0,0,1}{Senming Zhong}}
        \ead{itcsmzhong@gpnu.edu.cn}
	\address[3]{Industrial Training Center, Guangdong Polytechnic Normal University, Guangzhou, 510665, China }

	\cortext[cor1]{Corresponding author at: College Of Computer Science And Engineering,  Chongqing University of Technology, 400054, China} 



\begin{abstract}
Although current text detection algorithms demonstrate effectiveness in general scenarios, their performance declines when confronted with artistic-style text featuring complex structures. This paper proposes a method that utilizes Criss-Cross Attention and residual dense block to address the incomplete and misdiagnosis of artistic-style text detection by current algorithms. Specifically, our method mainly consists of a feature extraction backbone, a feature enhancement network, a multi-scale feature fusion module, and a boundary discrimination module. The feature enhancement network significantly enhances the model's perceptual capabilities in complex environments by fusing horizontal and vertical contextual information, allowing it to capture detailed features overlooked in artistic-style text. We incorporate residual dense block into the Feature Pyramid Network to suppress the effect of background noise during feature fusion. Aiming to omit the complex post-processing, we explore a boundary discrimination module that guides the correct generation of boundary proposals. Furthermore, given that movie poster titles often use stylized art fonts, we collected a \textbf{Movie-Poster} dataset to address the scarcity of artistic-style text data. Extensive experiments demonstrate that our proposed method performs superiorly on the Movie-Poster dataset and produces excellent results on multiple benchmark datasets. The code and the Movie-Poster dataset will be available at: \href{https://github.com/biedaxiaohua/Artistic-style-text-detection}{https://github.com/biedaxiaohua/Artistic-style-text-detection}
\end{abstract}


		

	\maketitle
\begin{figure*}[width=\textwidth,ht!]
    \centering
     \includegraphics[height=0.3\textwidth,width=\textwidth]{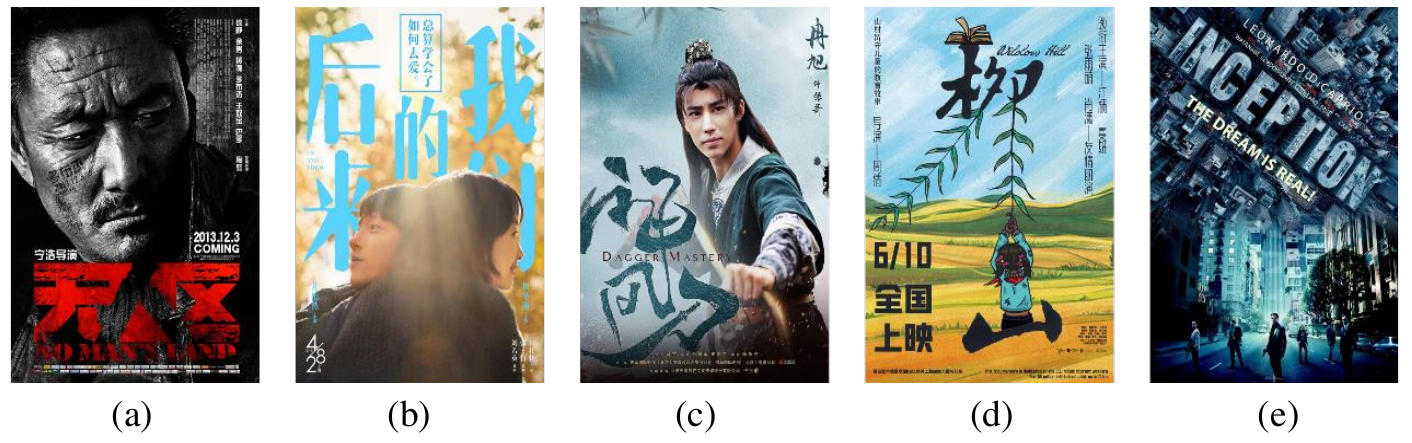}	    
     \caption{Some samples from the Movie-Poster dataset. Text in (a), (c), and (d) are highly artistic-style, text in (b) has widely varying aspect ratios, and in (e), the text area is mixed with background pixels.}
    \label{FIG_1}
\end{figure*}
\section{Introduction}\label{sec1}
\begin{figure*}[width=\textwidth,ht!]
    \centering
    \includegraphics[height=0.3\textwidth,width=1\textwidth]{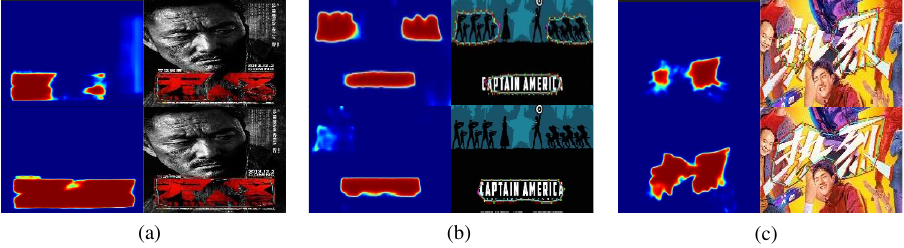}	
    \caption{Comparison of the effectiveness of our method with TextBPN++ on predicting mask maps and boundary proposals for the poster dataset. The visualization results of the TextBPN++ are on top of the image, and our method's results are on the bottom. (a) and (c) demonstrate that we have solved the problem of incomplete detection. (b) demonstrates that we have solved the problem of non-text pixels being mistaken for text pixels.}
    \label{FIG_2}
\end{figure*}

Text detection, the process of locating and extracting textual information from various surfaces such as images, videos, or 3D environments, is a fundamental component in computer vision and artificial intelligence \cite{long2021scene}. In the digital age, where multimedia content is abundant, accurate text detection is crucial to enhance search capabilities, automate data entry, and enable intelligent image understanding. Following the groundbreaking success of convolutional networks exemplified by AlexNet \cite{krizhevsky2017imagenet}, the field of text detection has experienced swift advancement. Several regression-based methods (\cite{tian2016detecting,liao2017textboxes,zhou2017east,wang2020contournet}) inspired by general object detection algorithms (\cite{girshick2015fast,fan2018salient,fan2019shifting}), employ rectangles or quadrilaterals for text instance localization. These methods have difficulties in detecting arbitrary shape text. To represent text instances flexibly, segmentation-based methods (\cite{deng2018pixellink,wang2019efficient,wang2019shape,liao2020real,zhu2021fourier}) treat text detection as a semantic segmentation task. These methods effectively facilitate text detection in general scenes. However, specific text environments pose unique challenges that require further research. As shown in Fig. \ref{FIG_1}, titles in movie posters are often artistically rendered to have a variety of shapes. As shown in Fig. \ref{FIG_2}, current algorithms have difficulty accurately recognizing and segmenting text regions, hindering effective feature learning. On the other hand, there is very little artistic-style text data on the market now. Therefore, detecting artistic-style text and supplementing artistic-style data is challenging and meaningful.\par
In existing datasets such as Total-Text \cite{ch2017total} and CTW1500 \cite{yuliang2017detecting}, the majority of text instances have a more standardized shape, with only a few artistic-style text included. Therefore, we collected the \textbf{Movie-Poster} dataset to enrich the currently scarce artistic-style text data. Specifically, this dataset contains 1500 movie posters, of which 1100 are used for training and 400 for testing. In our dataset, poster titles are often treated artistically. As shown in Fig. \ref{FIG_1}(a), (c), and (d), these titles are highly personalized and stylized; they can appear at any angle, and there may be connections and overlaps between characters. As shown in Fig. \ref{FIG_1}(b) and (e), some instances have significant differences in aspect ratios, and image background elements may be mixed with the text, making the detection of text regions more difficult.\par
This paper introduces a novel feature enhancement network, primarily composed of the \textbf{Recycle Criss-Cross Attention module (RCCA)}. This module allows for effective information transfer and interaction between feature maps of different channels, capturing semantic information about different objects, textures, and structures in an image to enhance the model's perceptual capabilities. It effectively addresses the issue of incomplete detection of shape-changing artistic-style text, as shown in Fig. \ref{FIG_2}(a) and (c).
TextPMS \cite{zhang2022arbitrary} and TextBPN++ \cite{zhang2023arbitrary} fuse the feature maps output by backbone through the Feature Pyramid Network (FPN), which effectively improves the representation of features. However, it still suffers from background noise interference for artistic-style text with complex backgrounds. Based on this, we design a new multi-scale feature fusion approach \textbf{Residual Feature Pyramid Network (R-FPN)}, combining residual dense structure and FPN to suppress the effect of redundant information effectively. As shown in Fig. \ref{FIG_2}(b), we have effectively solved the problem of incorrectly recognizing non-text pixels as text pixels. As a mixture of strokes and background pixels, text regions are highly homogeneous textures that do not have natural and well-defined boundaries. TextBPN++ \cite{zhang2023arbitrary} systematically presents a unified coarse-to-fine framework via boundary learning for arbitrary shape text detection. However, this method can lead to boundary modelling distortions for artistic-style text areas with extreme aspect ratios. To this end, we explore a \textbf{Boundary Discriminant Module (BDM)} that guides the generation of boundary proposals by combining the priori information with the feature maps output from the feature enhancement network. Subsequently, the boundary proposals are fed into the boundary transformer module for refinement, thus omitting a series of complex post-processing procedures.\par
In summary, our main contributions are four-fold:\par
\begin{itemize}
\item We propose a new feature enhancement network and feature fusion method that greatly enhances the model's perceptual capabilities, effectively mitigates the problem of incomplete detection and suppresses the effect of background noise.\par
\item We further explore a boundary discrimination module that guides the accurate generation of boundary proposals.\par
\item We propose the Movie-Poster dataset to complement the existing shortage of artistic-style text data.\par
\item Extensive experiments have demonstrated that our method achieves state-of-the-art performance on the Movie-Poster dataset and is competitive on the publicly available datasets.\par
\end{itemize}\par
The rest of the paper is organized as follows: Sect. \ref{Related_work} overviews the related work. Sect. \ref{Methodology} elaborates on our work. In Sect. \ref{sec4}, we demonstrate some experimental results and analysis. Sect. \ref{tab5} further discussed our proposed Movie-Poster dataset. Finally, we conclude our work in Sect. \ref{sec6}.
	\section{Related work}
	\label{Related_work}
With the rapid development of deep learning and object detection technology (\cite{kim2018pedestrian,kessentini2019two,jazayeri2019automatic}), significant progress has been made in the field of text detection \cite{liu2019scene}. Text detection can be broadly categorized into regression-based, segmentation-based, and connected component-based methods.\par
\subsection{Regression-based Methods}\label{subsec2}
The regression-based methods (\cite{wu2017self,zhou2017east,liao2017textboxes,ma2018arbitrary,wang2018geometry,xu2019textfield,zhu2021textmountain,zhong2022text}) rely on a regression-based object detection framework with word-level and line-level a priori knowledge. The difficulty of text detection is that text has irregular shapes with various aspect ratios, unlike standard images. RRPN \cite{ma2018arbitrary} and TextBoxes \cite{liao2017textboxes} effectively address this problem by predicting anchor offsets to localize text boxes. Other methods (\cite{zhou2017east,he2021most,he2017deep}) are anchor-free methods that directly regress the offset from a boundary or vertex to the current point. EAST \cite{zhou2017east} directly predicts text or lines of text in images with arbitrary orientations and rectangular shapes, eliminating unnecessary intermediate steps. SAST \cite{wang2019single} draws on the ideas of TextSnake \cite{long2018textsnake} and EAST \cite{zhou2017east} and joins the prediction of some geometric features of the text (the text centerline region, text boundary bias, and text center bias, etc.) while directly regressing on the bounding box so that it can be applied to irregular text detection. MOST \cite{he2021most} can dynamically adjust the receptive field of the localized prediction layer and adaptively merge the raw detections according to the predicted position, resulting in better detection of instances with large aspect ratios. TPLAANet \cite{zhong2022text} explores the prediction center mask to address the challenge of detecting text. Although the regression-based method is effective in quadrilateral text detection, it is not adapted to arbitrary shape text detection.
\subsection{Segmentation-based Methods}\label{subsec2}
Inspired by image segmentation methods (\cite{he2017mask,badrinarayanan2017segnet,schu2018new,yang2020automatic,farshi2020multimodal}), segmentation-based text detection methods (\cite{wang2019shape,liao2020real,liao2022real,li2024enhancing,yu2023turning}) classify pixels at the pixel level to discriminate whether each pixel point belongs to a text target and its connectivity with the surrounding pixels. Then, they integrate the results of neighbouring pixels into a text box. These methods can be adapted to any shape and angle of the text.  For example, PSENet \cite{wang2019shape} localizes text of any shape by pixel-level segmentation and uses a progressive scale expansion algorithm to identify neighbouring text instances. PAN \cite{wang2019efficient} improves on PSENet \cite{wang2019shape} by using a learnable post-processing method, pixel aggregation, to guide text pixels to correct kernel parameters by predicted similarity vectors and to reconstruct complete text instances from the predicted kernel to reconstruct complete text instances. LSAE \cite{tian2019learning} proposed pixel embedding, which groups pixels based on segmentation results to achieve more accurate text localization. DBNet \cite{liao2020real} embeds the threshold transformation process into the network for training by learning the threshold mapping and employing differentiable operations. In this way, the text detection model can adaptively learn the thresholds to capture the segmentation information of text more efficiently. DBNet++ \cite{liao2022real} adds the Adaptive Scale Fusion (ASF) module to DBNet \cite{liao2020real}. Features at different scales are processed through the ASF module to obtain better-fused features. CBNet \cite{zhao2024cbnet} uses knowledge distillation to detect arbitrary shape text efficiently. The most important thing for segmentation-based methods is to ensure the accuracy of segmentation. 
\begin{figure*}[width=\textwidth,ht!]\centering
     \includegraphics[height=0.5\textwidth,width=1\textwidth]{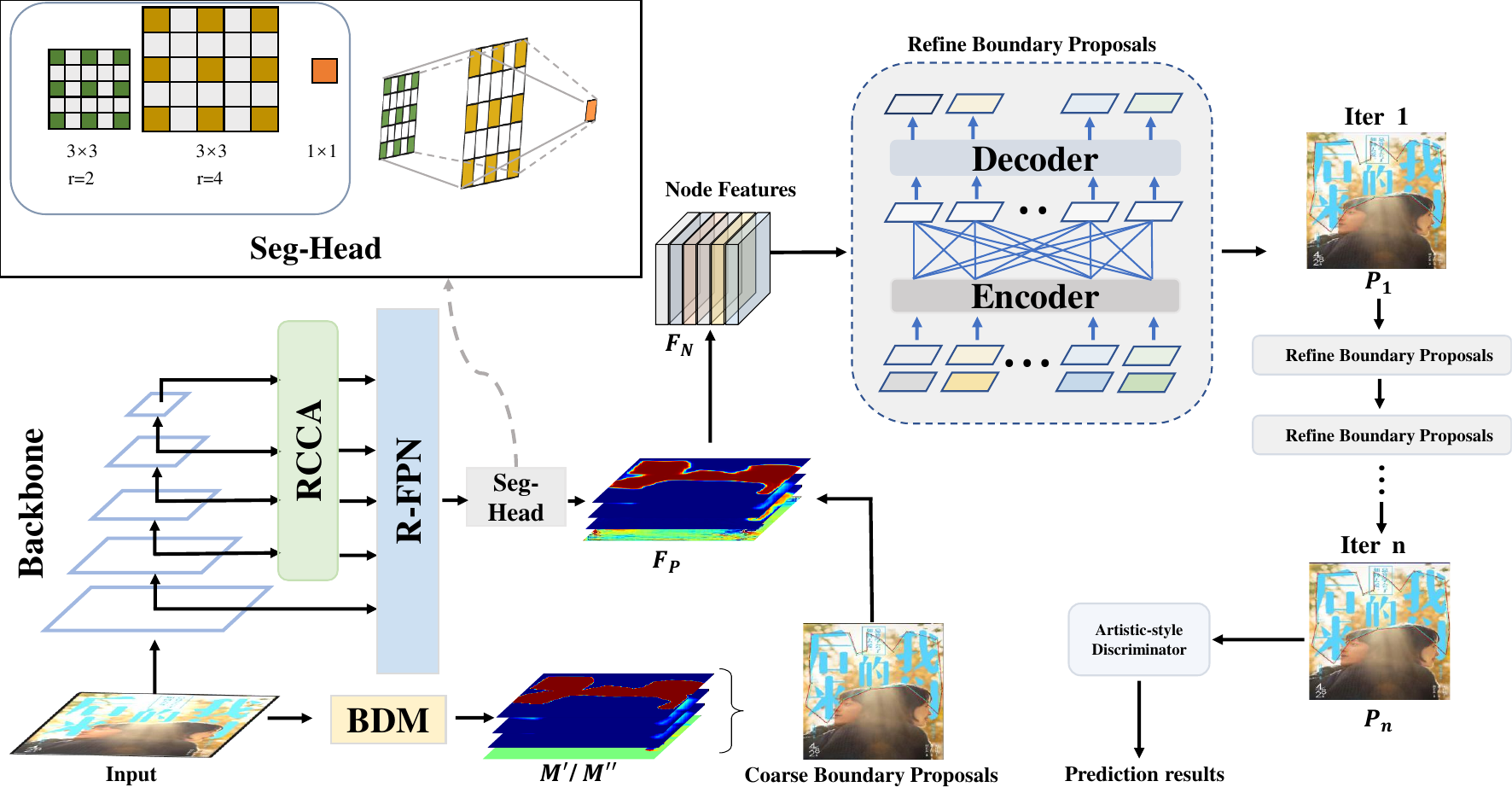}	    
\caption{Framework of our method. $F_{P}$ denotes the feature maps output by Seg-Head, $M^{'}/M^{''}$ denotes the masks generated by BDM, $F_{N}$ denotes the node features obtained by sampling on $F_{P}$, $P_{n}$ are the proposals after iteratively. We use the Artistic-style Discriminator to determine whether the predicted detection proposals contain non-artistic-style text. The Prediction results are the final predicted boundary proposals.}
\label{FIG_3}
\end{figure*}
\subsection{Connected Component-based Methods}\label{subsec2}
Connected Component-based Methods usually link or group the detected individual text parts or characters into final text instances through a post-processing process, for example (\cite{yin2015multi,shi2017detecting,baek2019character,feng2019textdragon,tang2019seglink++,zhang2020deep}). CRAFT \cite{baek2019character} innovatively breaks down the text detection task into two more recognizable components: fragments and links. In this framework, each text part is treated as separate \text{fragments}, while "links" connect two neighbouring fragments belonging to the same word, thus enabling complete detection of the entire word. SegLink \cite{shi2017detecting}, inspired by TextSnake \cite{long2018textsnake}, simply groups detected local text regions by their geometric relationships. DRRG \cite{zhang2020deep} uses Graph Convolutional Neural Networks \cite{kipf2016semi} (GCN) to learn and infer linking relationships of text components to group text components. Although linked component-based methods can work well for arbitrary shape text detection, they usually have complex post-processing procedures.

	\section{Methodology}
	\label{Methodology}
\subsection{Overall network architecture}\label{subsec2}
Our proposed model architecture is shown in Fig. \ref{FIG_3}. It consists of five main components: feature extraction backbone, feature enhancement network, feature fusion module, boundary discrimination module, and boundary transformer. Specifically, we adopt ResNet-50 \cite{he2016deep} as the backbone to extract features. We input the feature maps output from the backbone to the feature enhancement module, RCCA, which can accurately capture the semantic information in complex images more acutely. Further, we propose a novel feature fusion strategy, R-FPN, to integrate the multi-scale feature maps processed by feature enhancement. After that, the coarse boundary proposals generated by BDM are utilized to sample the feature maps and extract key node features. Finally, these features are fed into the boundary transformer for refinement to achieve more accurate text boundary localization.
\subsection{Recycle Criss-Cross Attention Module}\label{subsec2}
Current text detection algorithms often encounter issues with incomplete detection when dealing with artistic-style text. This is because they cannot effectively capture the complex structural information of artistic-style text. Therefore, we propose an RCCA module to enhance the model's perception of complex structures.\par
\begin{figure*}[width=\textwidth,ht!]\centering
     \includegraphics[height=0.3\textwidth,width=1\textwidth]{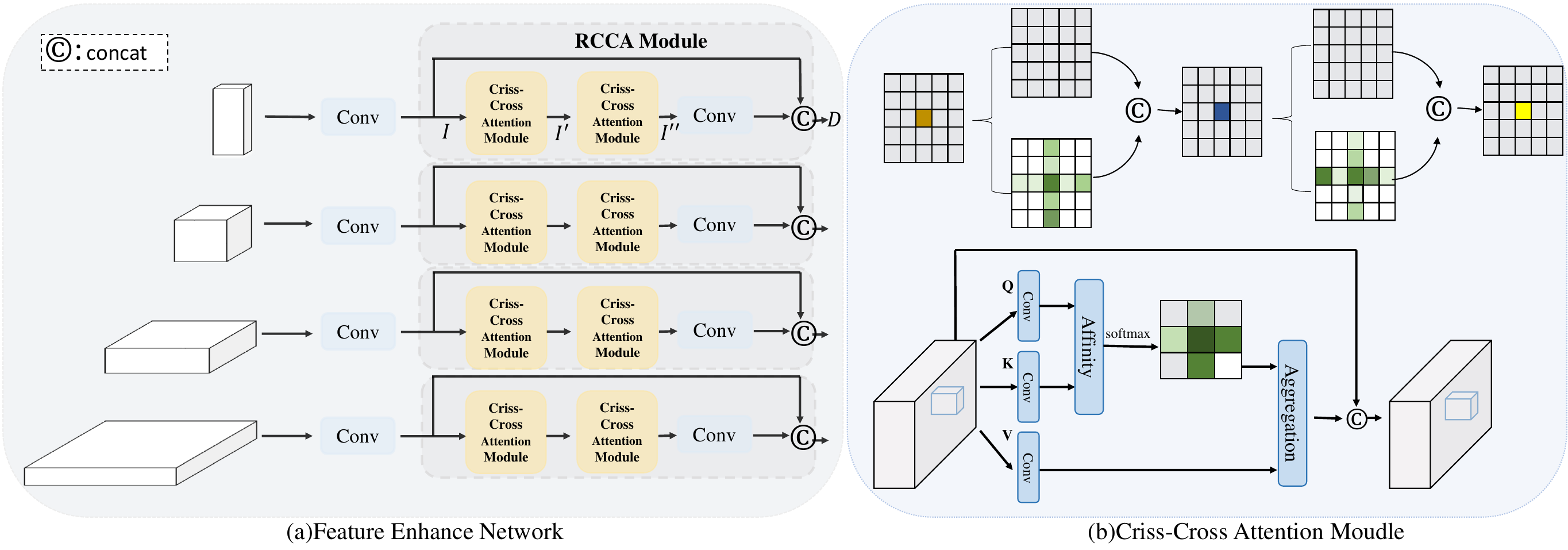}	    
     \caption{(a) The overall framework of the feature enhancement network includes four layers of the RCCA module. $I$, $I^{'}$, $I^{''}$, and $D$ denote the feature maps after various operations. (b) The Criss-Cross Attention Module framework.}
    \label{FIG_4}
\end{figure*}

The structure is shown in Fig. \ref{FIG_4}(a). First, we pass the feature map output from the backbone through a convolutional layer with 3×3 convolutional kernels for feature dimensionality reduction, which can effectively reduce redundant features and model computation. Next, we feed the feature maps into the Criss-Cross Attention Module \cite{huang2019ccnet}, which aggregates contextual information in the horizontal and vertical directions for each pixel, enhancing the model's perceptual capabilities. Then, through a loop operation, each pixel can eventually capture the global dependencies of all pixels. Specifically, after a layer of 3×3 convolution, our feature maps are dimensionalized down to 1/4 of the original ones, and then these feature maps are fed into the Criss-Cross Attention Module. The structure of the Criss-Cross Attention Module is shown at the top of Fig. \ref{FIG_4}(b). In the Criss-Cross Attention Module, the feature map $I\in {} R^{C\times W\times H} $ is firstly convolved by two 1×1 convolutions to generate $Q$ and $K$, where ${Q,K}\in R^{C^{'}\times W\times H }$, $C^{'}$ is the number of channel. After obtaining feature maps $Q$ and $K$, we further generate attention maps $A \in R^{(H+W-1) \times W \times H}$ via Affinity operation. The Affinity operation is defined as follows: \begin{eqnarray}
		d_{i,u}=Q_{u} \Upsilon _{i,u}
\end{eqnarray}
where $d_{i,u}$ is the degree of correlation between feature $Q_{u}$ and $\Upsilon {i,u}$. $Q_{u}\in R^{C^{'} }$ is the vector of $Q$ at $u$, $\Upsilon {u}$ are the feature vectors of $K$ which are in the same row or column with position $u$, $\Upsilon {i,u}$ is the vector of $\Upsilon _{u}$ at $i$. After getting all the $d_{i,u}$, a softmax operation is performed to the attention map $A$. Another convolutional layer with 1×1 convolutional kernels is applied on $I$ to generate $V\in R^{C\times W\times H }$ for feature adaption. Similarly, at the same position $u$ of $V$, we get the feature vectors $V_{u} \in R^{C}$, and finally, we get a set $\Psi _{u} \in R^{(H+W-1)\times C}$, which is the set extracted from the same rows and columns as $u$ in $V$. Further, we get the semantic information of the duplicate rows and columns at position u through the aggregation operation, which is defined as follows:
\begin{eqnarray}
		I_{u}^{'} =\sum_{i\in \left | \Psi _{u}  \right | }^{}A_{i,u}\Psi _{i,u}+I_{u} 
\end{eqnarray}
where $I_{u}^{'}$ is a feature vector in output maps $I^{'}\in R^{C\times W\times H }$. $A_{i,u}$ is a scalar value at channel $i$ and position $u$ in $A$. The contextual information is added to
local feature $I$ to enhance the local features and augment
the pixel-wise representation.\par
As shown at the bottom of Fig. \ref{FIG_4}(b), since the remote context information is only captured horizontally and vertically in the Criss-Cross Attention Module, and the pixel-to-pixel connections around the pixel are still sparse, we designed a recycle module where the output feature map $I^{'}$ after passing through the first module is fed to the next Criss-Cross Attention Module so that our feature maps $I^{''}$ contain global context information. The operation is defined as follows:
\begin{eqnarray}
		I^{''}=Att(I^{'})   
\end{eqnarray}
where $Att$ denotes the operation of Criss-Cross Attention, $I^{''}\in R^{C\times W\times H }$ is the feature map output after two $Att$ operations.\par
Then $I^{''}$ again undergo a convolution operation and concat with $I$ to get the final output feature maps $D\in R^{C\times W\times H }$.
\subsection{Residual Feature Pyramid Network}\label{subsec2}
\begin{figure}[ht!]\centering
     \includegraphics[height=0.75\textwidth,width=0.45\textwidth]{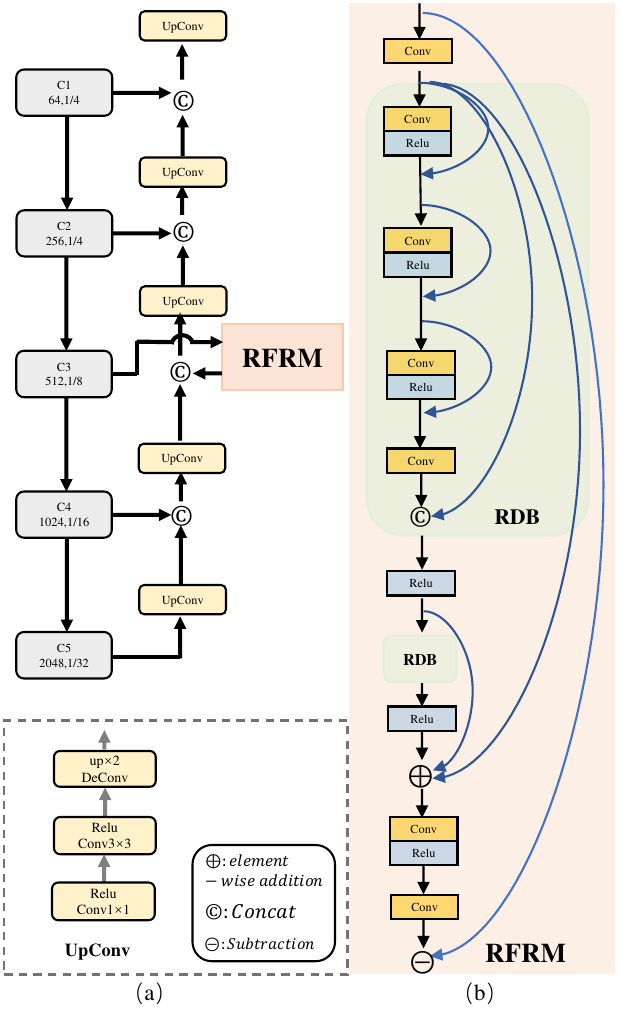}	    
     \caption{(a) The overall structure of R-FPN. The 1/64, 1/256, 1/1024, and 1/2048 indicate the scale ratio compared to the input image. (b) Detailed architecture of RFRM in R-FPN.}
    \label{FIG_5}
\end{figure}

Since artistic-style text is often mixed with background pixels, the current method usually mistakes the background for the text. Therefore, we propose an FPN-based feature fusion method, R-FPN, to suppress the effect of background noise. Its structure is shown in Fig. \ref{FIG_5}(a).\par
Specifically, we change the input at the third layer of the fusion stage and feed the feature map $D\in R^{C\times W\times H }$ output from the feature enhancement network into our proposed Redundant Feature Reduction Module (RFRM) to obtain a clean feature map $D^{'}\in R^{C\times W\times H }$, which is then fed into the feature pyramid network for multiscale fusion. The structure of the RFRM is shown in Fig. \ref{FIG_5}(b). In RFRM, we first pass a feature map of size 512×512 through a convolutional layer with 3×3 convolutional kernels to better capture the global information. Then, we use the Residual Dense Block (RDB) with a four-layer structure to get the background noise features $O\in R^{C^{'} \times W \times H}$. The RDB consists of residual structure and ReLU operation. It extracts rich local features through densely connected convolutional layers and introduces local residual learning to improve the information flow further. The residual operation is next used to fuse the features obtained from the RDB and ReLU combination, effectively preventing the long-term dependency problem. After the second RDB module and the ReLU operation, we perform an element-wise addition operation on the input raw features and the features from the first and second RDB block to get the final redundant features $O^{'}\in R^{C^{'} \times W \times H}$. Finally, we subtract the redundant features obtained by RFRM from the original input features to obtain a clean feature map $D^{'}$, and then input $D^{'}\in R^{C\times W\times H }$ into FPN for feature fusion, which is defined by RFRM as follows:
\begin{eqnarray}
D^{'}=D-CRC(O+O^{'} )
\end{eqnarray}
where $D^{'}\in R^{C\times W\times H }$ is the feature map after denoising, $C$ denotes a function of Conv, $R$ express a function of ReLU.

\subsection{Boundary Discrimination Module}\label{subsec2}
Due to irregular shapes and extreme aspect ratios, artistic-style text can suffer from boundary modelling distortion. In this regard, we propose the BDM to solve this problem.\par
Specifically, we map ground truth onto the input image to obtain a mask map $M\in R^{C \times W \times H}$. Based on $M$, we compute the Euclidean Distance transform from each pixel point in the image to the nearest nonzero pixel point to obtain a new mask map $M^{'}\in R^{C \times W \times H}$. From the output mask map $M^{'}$, we can compute the coarse boundary proposals $P_{0}$. However, the initially generated coarse boundary proposal may be erroneous due to some images' highly irregular shapes of artistic-style text regions. This will introduce noise during training and affect the model's performance. Therefore, we design a discriminator to solve this problem. If the number of nonzero pixels in $M^{'}$ is much lower or higher than the number of nonzero pixels in $M$, it is determined that its modelling fails. If it fails, the Euclidean Distance transform between each pixel point in the ground truth mask and the nearest nonzero pixel point is computed to obtain the mask map $M^{''}$, and then the coarse boundary proposals are generated based on $M^{''}$.
\begin{figure*}[width=\textwidth,ht!]\centering
     \includegraphics[height=0.34\textwidth,width=1\textwidth]{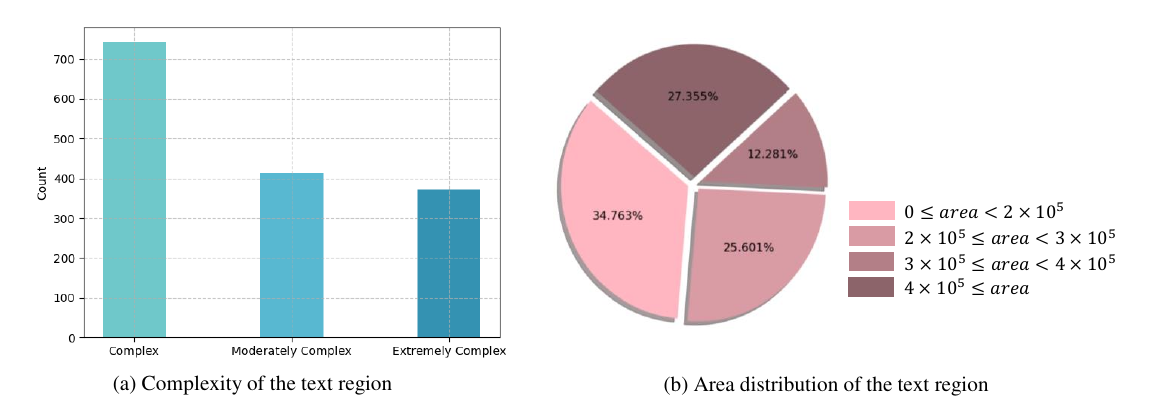}	    
     \caption{Some analytical visualizations of the Movie-Poster dataset. (a) Complexity of the text region in Movie-Poster. We judge the complexity of a text region by the number of coordinate points, with coordinate points between 10 and 15 being Complex, coordinate points between 15 and 30 being Moderately Complex, and coordinate points than 30 being Extremely Complex. (b) Area distribution of all text regions in Movie-Poster, where the unit is square pixels.}
    \label{FIG_6}
\end{figure*}
\begin{equation} 
P_{0}= \left\{  
             \begin{array}{lr}  
             f(M^{''}) ,& \frac{N_{M^{'} } }{N_{M}}<th_{a}\    or\  \frac{N_{M^{'} }}{N_{M}}>th_{b}     \\  
             f(M^{'} ), & th_{a}<=\frac{N_{M^{'} } }{N_{M}}<=th_{b}\\  

             \end{array}  
\right.
\end{equation} 
where  $f$ stands for the function of calculates the coarse boundary proposals based on the mask map; $N_{M}$, $N_{M^{'}}$ denotes the number of nonzero pixels in $M$ and $M^{'}$; $th_{a}$ and $th_{b}$ is the threshold that is experimentally proven to perform best when $th_{a}=0.25$ and $th_{b}=1.75$;\par
The Seg-Head consists of a multilayer expansion convolution, including two 3 × 3 convolution layers with different expansion rates and one 1 × 1 convolution layer. With the Seg-Head, we obtain the feature maps $F_{P}\in R^{C \times W \times H}$: the classification map, the distance field map, and the direction field map. We root $P_{0}$ to sample the specified location of the feature maps $F_{P}$ to obtain the node features $F_{N}$. Then $F_{N}$ are fed into the boundary transformer for n iterations of refinement to obtain the final boundary proposals $P_{n}$.

\begin{table*}[width=\textwidth]
\centering
\renewcommand\arraystretch{1.2}
\caption{Ablation study of R-FPN, RCCA, BDM and Artistic-style Discriminator. '$\checkmark$' means use the module, and ‘$\times$’ means don't use it. \textbf{Bold} indicates the best result, while \underline{underline} indicates second.}
\label{tab1}
\begin{tabular}{ccccccccccl}
\hline
\multicolumn{4}{c}{Methods}     & \multicolumn{3}{c}{Movie-Poster} & \multicolumn{4}{c}{Total-Text}                         \\
R-FPN & RDM & RCCA &  Discriminator & Precision   & Recall   & F-measure  & Precision   & Recall   & \multicolumn{2}{c}{F-measure} \\ \hline
$\times$     & $\times$   & $\times$    & $\times$          & 71.92       & 86.83    & 78.67      & \underline{91.1}        & 83.4     & \multicolumn{2}{c}{87.08}     \\
$\checkmark$     & $\times$   & $\times$    & $\times$          & 81.48       & 85.85    & 83.61      & 90.4        & 84.3     & \multicolumn{2}{c}{87.26}     \\
$\times$     & $\checkmark$   & $\times$    & $\times$          & 84.67       & 84.87    & 84.77      &\textbf{91.12}             &83.4          & \multicolumn{2}{c}{87.09}          \\
$\times$     & $\times$   & $\checkmark$    & $\times$          & \underline{86.84}       & 83.66    & 85.22      & 90.5            & 84.86         & \multicolumn{2}{c}{87.59}          \\
$\checkmark$     & $\times$   & $\checkmark$    & $\times$          & 81.95       & \textbf{89.76}    & 85.68      & 90.27           & \underline{85.53}         & \multicolumn{2}{c}{\underline{87.84}}          \\
$\checkmark$     & $\checkmark$   & $\checkmark$    & $\times$          & 83.76       & \underline{88.05}    & \underline{85.85}      & 90.17       & \textbf{85.67}    & \multicolumn{2}{c}{\textbf{87.86}}     \\
$\checkmark$     & $\checkmark$   & $\checkmark$    & $\checkmark$          & \textbf{88.89}       & 85.85    & \textbf{87.34}      & -           &-         & \multicolumn{2}{c}{-}          \\ \hline
\end{tabular}
\end{table*}
\begin{table}[]
\renewcommand\arraystretch{1.2}
\caption{Ablation study of the number of Criss-Cross Attention Module cycles. 'P', 'R', and 'F' represent Precision, Recall, and F-measure. \textbf{Bold} indicates the best result, while \underline{underline} indicates second.}
\label{tab2}
\begin{tabular}{cccc}
\hline
\multicolumn{1}{l}{} & \multicolumn{3}{c}{Movie-Poster} \\
Cycles num.          & Precision  & Recall & F-measure \\ \hline
0                    & 81.5       & \textbf{85.89}   & 83.64     \\
1                    & 87.28      & 83.66   & 85.43     \\
2                    & \textbf{88.89}      & \underline{85.85}   & \textbf{87.43}     \\
3                    & \underline{88.49}      & 84.39   & \underline{86.39}     \\ \hline
\end{tabular}
\end{table}
\section{Experimental results}\label{sec4}
\subsection{Datasets}\label{subsec2}
\textbf{Movie-Poster:} It contains 1500 images, of which 1100 are in the training set and 400 in the testing set. The data consists of movie posters, most of which are artistic-style text of arbitrary shapes. These text regions can appear at any angle, and there may be joins and overlaps between characters. Fig. \ref{FIG_6} illustrates some of the analysis of the Movie-Poster dataset. This shows that our dataset contains text boxes of various shapes and sizes, which is highly challenging. In addition, the Movie-Poster dataset is multilingual, mainly in English and Chinese.\par
\textbf{Total-Text (\cite{ch2017total}):} It is a dataset containing the text of various shapes, including horizontal, multi-orientational, and curved. The dataset has a total of 1555 images and 11459 text lines. There are 1255 images in the training set and 300 images in the testing set.\par
\textbf{CTW1500 (\cite{yuliang2017detecting}):} It dataset contains 1,500 images, of which 1,000 are for training and 500 are for testing. Each image has at least one curved text, containing many horizontal and multi-orientational text. In addition, the dataset is multilingual, with mainly Chinese and English text.\par
\textbf{MSRA-TD500 (\cite{yao2012detecting}):} It contains 500 natural scene images, of which 300 are for training and 200 are for testing. It supports multiple languages, mainly Chinese and English. It contains a variety of text orientations and covers diverse scenes, such as billboards, signs, and walls.\par
\textbf{ICDAR-Art (\cite{chng2019icdar2019}):} It contains a total of 10,166 images, 5603 images in the training set and 4563 images in the testing set. It consists of three parts: Total-Text, SCUT-CTW1500, and Baidu Curved Scene Text (ICDAR2019-LSVT partially curved data), and contains horizontal, multi-orientational, and curved text of various shapes.
\subsection{Implementation Details}\label{subsec2}
The ResNet-50 \cite{he2016deep} is adopted as our backbone. In our experiments, we randomly crop the text region and resize it to
640 × 640 for training the model. To demonstrate the model's performance, we do not use pre-training data, and the initial learning rate is set to 0.0001 or 0.001 and decays by 0.9 after every 50 epochs. For all datasets, the training batch size is set to 4, and we use Adam \cite{kingma2014adam} as the optimizer. The data augmentation includes random rotation with an angle, random cropping, and random flipping. The other datasets were trained with 600 epochs except for MSRA-TD500, which was trained with 1200 epochs. In the inference, we keep the aspect ratio of the test images and then resize and fill them to the same size for the test. Training and testing are performed on a single GPU (NVIDIA GeForce RTX3090).
\begin{table*}[width=\textwidth]
\centering
\renewcommand\arraystretch{1.2}
\caption{Detection results on the Movie-Poster dataset. \textbf{Bold} indicates the best result, while \underline{underline} indicates second. $^{\dag}$ denotes the result reproduced using the original paper method.}
\begin{tabular}{cccccccc}
\hline
Methods         & Published   & \multicolumn{3}{c}{Movie-Poster/IOU 0.5} & \multicolumn{3}{c}{Movie-Poster/IOU 0.75} \\
                &             & Precision      & Recall     & F-measure     & Precision      & Recall      & F-measure     \\ \hline
PSENet$^{\dag}$\cite{wang2019shape}          & CVPR'19     & 79.54          & 84.39      & 81.89         & 60.68          & 64.39       & 62.48         \\
FAST$^{\dag}$\cite{chen2021fast}            & Arxiv'21. & 74.65          & 78.29      & 76.42         & 46.38          & 48.53       & 47.43         \\
PAN$^{\dag}$\cite{wang2019efficient}             & ICCV’19     & 82.23          & 82.43      & 82.33         & 63.26          & 63.41       & 63.33         \\
DBNet++$^{\dag}$\cite{liao2022real}         & TPAMI’22    & 80.2           & 77.45      & 78.8          & -              & -           & -             \\
TextPMS$^{\dag}$\cite{zhang2022arbitrary}         & TPAMI’23    & 75.75          & \underline{86.09}      & 80.59         & 62.01          & \textbf{70.48}       & 65.98         \\
TextBPN++$^{\dag}$\cite{zhang2023arbitrary}       & T-MM'23     & 71.92          & \textbf{86.83}      & 78.67         & 44.84          & 54.14       & 49.06         \\
CBNet$^{\dag}$\cite{zhao2024cbnet}           & IJCV'24     & \underline{87.89}          & 81.46      & \underline{84.55}         & \underline{68.68}          & 63.65       & \underline{66.07}         \\
Ours & -           & \textbf{88.89}          & 85.85      & \textbf{87.34}         & \textbf{70.7}           & \underline{68.29}       & \textbf{69.47}         \\ \hline
\label{tab3}
\end{tabular}
\end{table*}
\subsection{Ablation study}\label{subsec2}
We perform a series of ablation experiments on the Movie-Poster dataset and Total-Text to demonstrate the effectiveness of our proposed modules. A detailed study of the performance of each module is given in Table \ref{tab1}.\par
\textbf{Effectiveness of RCCA: }Table \ref{tab1} demonstrates the effectiveness of the RCCA module. On the Movie-Poster dataset, the use of the RCCA module alone results in a significant increase in F-measure by about 6.55$\%$. This proves that RCCA can effectively utilize the global semantic information to better adapt to complex environments, significantly improving the model's effectiveness in detecting artistic-style text. In addition, for the Total-Text dataset, our F-measure is also increased by 0.51$\%$, proving that the module is useful for detecting curved text.\par
\textbf{Effectiveness of R-FPN: }The effectiveness of the R-FPN can be seen from the results in Table \ref{tab1}. Only using this module on the Movie-Poster and Total-Text datasets improves the F-measure by 4.94$\%$ and 0.18$\%$, respectively. If both R-FPN and RCCA are used, the increment of F-measure reaches 0.78$\%$ on the totaltext and 7.01$\%$ on the Movie-Poster.\par
\textbf{Effectiveness of BDM: }BDM is also effective on the Artistic Poster dataset, as shown in Table \ref{tab1}. If only BDM is used, the increment of F-measure reaches 6.1$\%$. This shows that the correctness of boundary modelling is essential for detecting artistic-style text, and BDM effectively guarantees the correctness of boundary modelling. Since the text instances in the Total-Text have a more regular shape, BDM  has relatively little impact on this dataset, with a 0.01$\%$ increase in F-measure. Our method uses R-FPN, RCCA, and BDM, increasing F-measure by 0.78$\%$ on Total-Text and 7.18$\%$ on Movie-Poster.\par
\textbf{Effectiveness of the number of Criss-Cross Attention Module cycles: }As shown in Table \ref{tab2}, the detection effect is optimized when the number of cycles is set to 2. The analysis yields that the Criss-Cross Attention Module collects semantic information in each cycle's horizontal and vertical directions. When the number of cycles is 1, the connection between pixels is still sparse. Each pixel incorporates global semantic information by increasing the number of cycles to 2. However, if the number of cycles is increased to 3 or more, too much noise may be introduced, which leads to a decrease in the recall. Therefore, our method sets the number of cycles to 2 to balance the information's comprehensiveness with noise control.\par
When detecting the Movie-Poster dataset, we designed an Artistic-style Discriminator to determine whether the detected text region is an artistic-style title, effectively removing those redundant text boxes. The F-measure increases by 1.49$\%$, reaching 87.34$\%$, and the results are shown in Table \ref{tab1}.
\begin{figure*}[width=\textwidth,ht!]\centering
     \includegraphics[height=\textwidth,width=\textwidth]{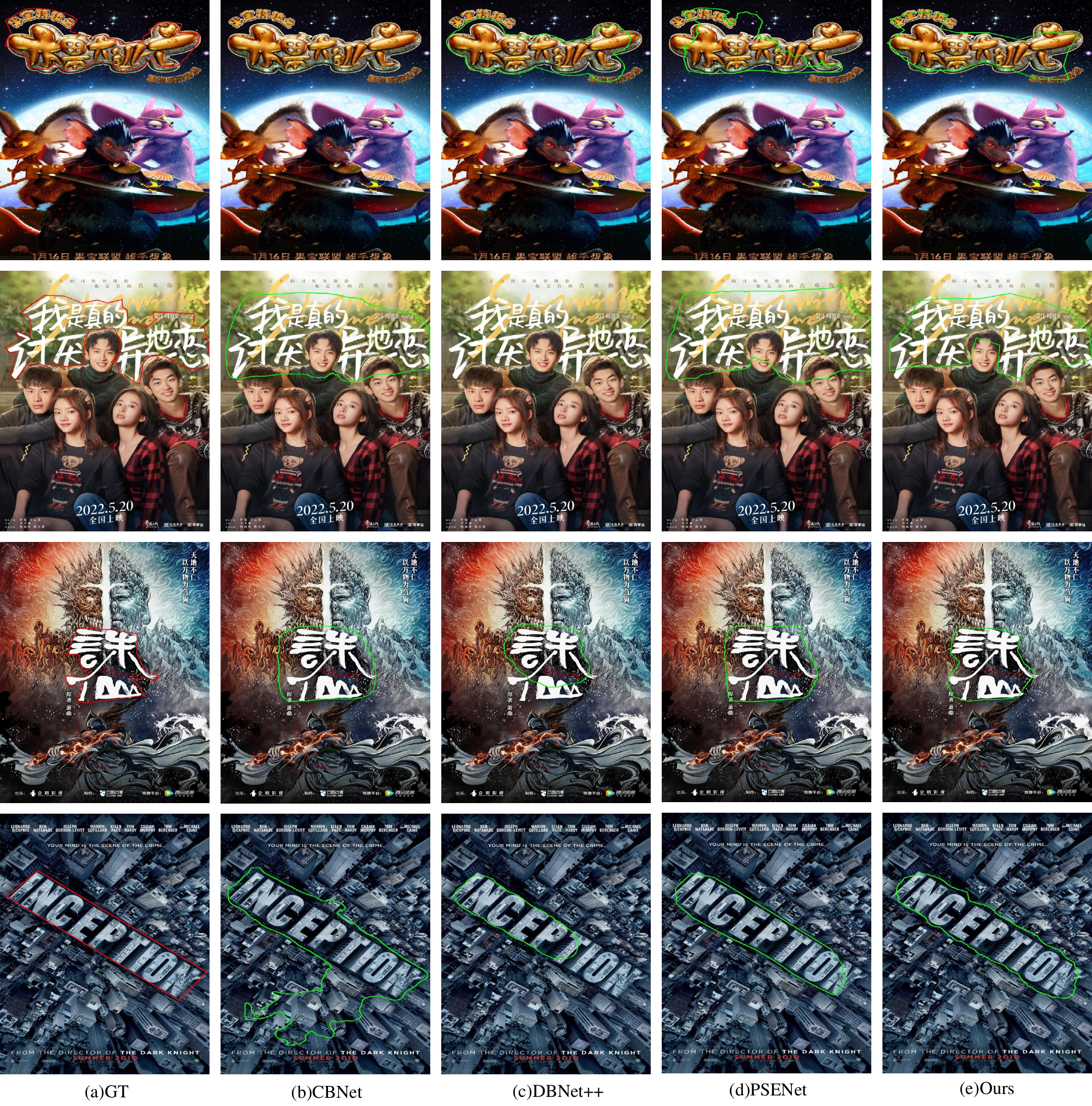}	    
     \caption{Qualitative comparisons with CBNet\cite{zhao2024cbnet}, DBNet++\cite{liao2022real}, and PSENet\cite{wang2019shape}.}
    \label{FIG_7}
\end{figure*}
\subsection{Comparisons with previous methods}\label{subsec2}
To comprehensively evaluate the performance of our proposed method, we conducted comparative experiments on five different datasets. These datasets include the Movie-Poster dataset and four recognized benchmark datasets: Total-Text, CTW1500, ICDAR-Art, and MSRA-TD500. For a comprehensive evaluation, we employed both qualitative and quantitative methods.\par
\textbf{Movie-Poster:} During training, we configured the input image dimensions to 640 × 640 and employed the Adam \cite{kingma2014adam} optimizer. We commenced the training process without leveraging any pre-trained data, setting the initial learning rate to 0.0001 and applying an exponential decay of 0.9 every 50 epochs. Given that the artistic-style text in the Movie-Poster dataset contains a variety of glyphs of larger sizes, we set intersection and concurrency (IOU) thresholds of 0.5 and 0.75  respectively when testing. The quantitative results are shown in Table \ref{tab3}. Our method achieved promising results in Precision (88.89$\%$), Recall (85.85$\%$) and F-measure (87.34$\%$). The qualitative results are shown in Fig. {\ref{FIG_7}}. Compared with the state-of-the-art methods (\cite{liao2022real,zhang2022arbitrary,zhang2023arbitrary,zhao2024cbnet}), we have a clear advantage, and all indicators have achieved the most advanced results. When evaluating the  F-measure, our method improves the performance by 8.67$\%$ compared to the baseline \cite{zhang2023arbitrary} when the intersection and concurrency ratio (IOU) threshold is set to 0.5 and improves the performance by 2.79$\%$ compared to the current state-of-the-art method CBNet \cite{zhao2024cbnet}. When the IOU threshold is increased to 0.75, our method further improves the performance by 20.41$\%$ compared to the baseline \cite{zhang2023arbitrary}, and 3.4$\%$ compared to the CBNet \cite{zhao2024cbnet}. This shows the effectiveness of our method for artistic-style text detection.\par
\begin{table}[]
\renewcommand\arraystretch{1.2}
\caption{Detection results on the Total-Text dataset.  'P', 'R', and 'F' represent Precision, Recall, and F-measure. \textbf{Bold} indicates the best result, while \underline{underline} indicates second. $^{\dag}$ denotes the result reproduced using the original paper method.}
\label{tab4}
\scalebox{0.8}{
\begin{tabular}{ccccccccccl}
\hline
Methods    & Published   & P     & R     & F     \\ \hline
SPCNet\cite{xie2019scene}     & AAAI'19     & 83    & 82.8  & 82.9  \\
LOMO\cite{zhang2019look}       & CVPR'19     & 87.6  & 79.3  & 83.3  \\
PSENet$^{\dag}$\cite{wang2019shape}     & CVPR'19     & 88.55 & 77.81 & 82.83 \\
PAN$^{\dag}$\cite{wang2019efficient}        & ICCV'19     & 83.54 & 77.45 & 80.38 \\
ContourNet\cite{wang2020contournet} & CVPR'20     & 86.9  & 83.9  & 85.4  \\
DRRG\cite{zhang2020deep}       & CVPR'20     & 86.5  & 84.9  & 85.7  \\
PAN++$^{\dag}$\cite{wang2021pan++}      & TPAMI’21    & 76.21 & 69.05 & 72.46 \\
FAST$^{\dag}$\cite{chen2021fast}       & Arxiv'21 & 87.44 & 79.29 & 83.17 \\
DBNet++$^{\dag}$\cite{liao2022real}    & TPAMI’22    & 87.48 & 78.94 & 82.3  \\
TextPMS$^{\dag}$\cite{zhang2022arbitrary}    & TPAMI’22    & 85.66 & 83.28 & 84.55 \\
EMA\cite{zhao2022mixed}        & TIP'22      & 83.3  & \textbf{88.9}  & 86    \\
TextBPN++$^{\dag}$\cite{zhang2023arbitrary}  & T-MM'23     & \textbf{91.1}  & 83.4  & \underline{87.08} \\
CBNet$^{\dag}$\cite{zhao2024cbnet}      & IJCV'24     & 87.53 & 80.21 & 83.71 \\
Ours       & -           & \underline{90.17} & \underline{85.67} & \textbf{87.86} \\ \hline
\end{tabular}}
\end{table}
\begin{table}[]
\renewcommand\arraystretch{1.2}
\caption{Detection results on the CTW1500 dataset. 'P', 'R', and 'F' represent Precision, Recall, and F-measure. \textbf{Bold} indicates the best result, while \underline{underline} indicates second. $^{\dag}$ denotes the result reproduced using the original paper method.}
\label{tab5}
\scalebox{0.8}{
\begin{tabular}{ccccc}
\hline
Methods   & Published   & P     & R     & F     \\ \hline
PSENet$^{\dag}$\cite{wang2019shape}    & CVPR'19     & 82.06 & 77.97 & 79.96 \\
LOMO\cite{zhang2019look}      & CVPR'19     & \textbf{89.2}  & 69.6  & 78.4  \\
PAN$^{\dag}$\cite{wang2019efficient}       & ICCV'19     & 78.18 & 78.85 & 78.51 \\
TextRay\cite{wang2020textray}   & MM'20       & 77.9  & 83.5  & 80.6  \\
PAN++$^{\dag}$\cite{wang2021pan++}     & TPAMI'21    & 74.18 & 74.8  & 74.49 \\
TextPMS$^{\dag}$\cite{zhang2022arbitrary}   & TPAMI'22    & 84.02 & 78.32 & 81.07 \\
FAST$^{\dag}$\cite{chen2021fast}      & Arxiv'21 & 83.51 & 76.27 & 79.73 \\
TextBPN++$^{\dag}$\cite{zhang2023arbitrary} & T-MM'23     & \underline{87}    & \underline{79.63} & \textbf{83.15} \\
Ours      & -           & 81.59  & \textbf{83.8}  & \underline{83.12} \\ \hline

\end{tabular}}
\end{table}
\begin{table}[]
\renewcommand\arraystretch{1.2}
\caption{Detection results on the MSRA-TD500 dataset. 'P', 'R', and 'F' represent Precision, Recall, and F-measure. \textbf{Bold} indicates the best result, while \underline{underline} indicates second. $^{\dag}$ denotes the result reproduced using the original paper method.}
\label{tab6}
\scalebox{0.8}{
\begin{tabular}{ccccc}
\hline
Methods   & Published   & P     & R     & F     \\ \hline
PAN$^{\dag}$\cite{wang2019efficient}       & ICCV'19     & 58.7  & 68.03 & 63.02 \\
TextPMS$^{\dag}$\cite{zhang2022arbitrary}   & TPAMI’22    & 69.79 & 74.23 & 71.94 \\
FAST$^{\dag}$\cite{chen2021fast}      & Chen et al. & \textbf{75.42} & 68.59 & 72.04 \\
TextBPN++$^{\dag}$\cite{zhang2023arbitrary} & T-MM'23     & 72.17 & 74.4  & 73.27 \\
CBNet$^{\dag}$\cite{zhao2024cbnet} & IJCV'24     & 75.13 & \textbf{76.79} & \textbf{75.95} \\
Ours      & -            & \underline{75.38} & \underline{76} & \underline{75.8} \\ \hline

\end{tabular}}
\end{table}
\begin{table}[]
\renewcommand\arraystretch{1.2}
\caption{Detection results on the ICDAR-Art dataset. 'P', 'R', and 'F' represent Precision, Recall, and F-measure. \textbf{Bold} indicates the best result, while \underline{underline} indicates second. $^{\dag}$ denotes the result reproduced using the original paper method.}
\label{tab7}
\scalebox{0.8}{
\begin{tabular}{ccccc}
\hline
Methods     & Published & P    & R     & F     \\ \hline
PSENet$^{\dag}$\cite{wang2019shape}      & CVPR'19   & \textbf{81.1} & 57.5  & 67.3  \\
ContourNet\cite{wang2020contournet}  & CVPR'20   & 62.1 & 73.2  & 67.2  \\
DBNet\cite{liao2020real}       & AAAI'20   & 56   & 69.9  & 62.2  \\
TextRay\cite{wang2020textray}     & MM'20     & 58.6 & 75.97 & 66.17 \\
PCR\cite{dai2021progressive}         & CVPR'21   & 65   & \textbf{83.6}  & 73.1  \\
EMA\cite{zhao2022mixed}         & TIP'22    & 68.7 & \underline{80.8}  & \underline{74.3}  \\
Wang et al.$^{\dag}$\cite{wang2022fuzzy} & TIP'23    & 60.6 & 78.35 & 68.4  \\
Ours        &-           & \underline{70.5} & 80.37 & \textbf{75.11} \\ \hline
\end{tabular}}
\end{table}
\textbf{Total-Text:} During training, we initialize the learning rate to 0.001, and other parameters are the same as training the Movie-Poster dataset. The quantitative results are shown in Table \ref{tab4}. Regarding the F-measure, our proposed method demonstrates a 0.78$\%$ enhancement in performance over the baseline \cite{zhang2023arbitrary} and a notable 4.15$\%$ improvement when compared with CBNet \cite{zhao2024cbnet}. The experimental outcomes provide compelling evidence of the efficacy of our methodology in detecting and processing curved text.\par
\begin{figure*}[width=\textwidth,ht!]
\centering
     \includegraphics[height=0.27\textwidth,width=1\textwidth]{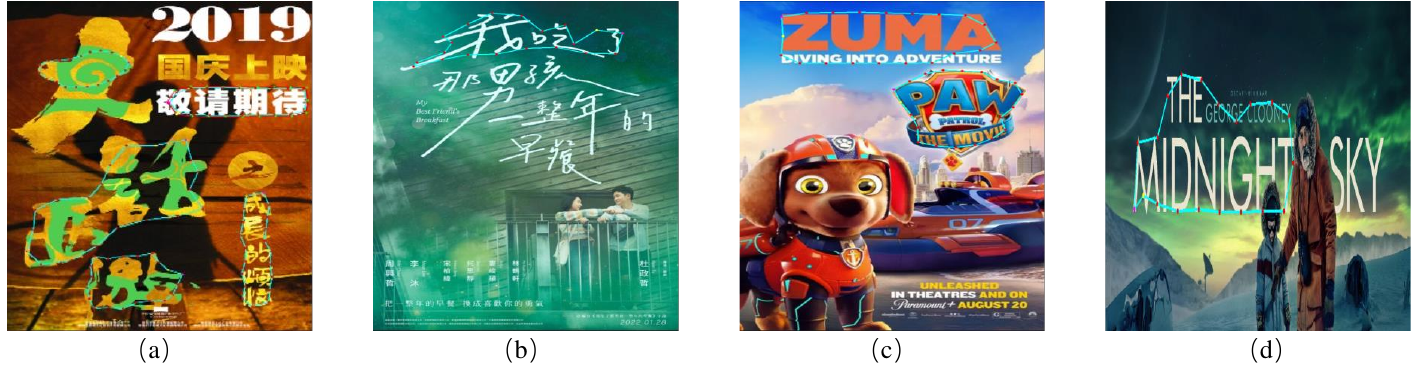}	    
     \caption{Some visual results of failure cases. In (a) and (b), these text regions are very large in shape and more similar in color to the background. In (c), our method does not exclude the redundant text boxes in this poster. In (d), the detection fails because a character image splits the overall text region.}
    \label{FIG_8}
\end{figure*}
\textbf{CTW1500:} Similarly, during training, we configured the input image dimensions to 640 × 640 and employed the Adam \cite{kingma2014adam} optimizer. We commenced the training process without leveraging any pre-trained data, setting the initial learning rate to 0.0001 and applying an exponential decay of 0.9 every 50 epochs. As shown in Table \ref{tab5}, the quantitative analysis results on the CTW1500 dataset indicate that our proposed method exhibits similar performance to TextBPN++ \cite{zhang2023arbitrary}. Further comparing with method TextPMS \cite{zhang2022arbitrary}, our F-measure achieves a 2.05$\%$ improvement.\par
\textbf{MSRA-TD500:} Considering the limited amount of data in the dataset MSRA-TD500, we performed 1200 epochs during the training process and set the initial learning rate to 0.0001. In the testing stage, we adjusted the size of the input image to limit it to 640 × 960 pixels. As illustrated in the Table \ref{tab6}, our proposed method achieves a 2.6$\%$ improvement in the F-measure compared to the TextBPN++\cite{zhang2023arbitrary}. In addition, we are also competitive with the state-of-the-art method CBNet\cite{zhao2024cbnet}.\par
\textbf{ICDAR-Art:} To substantiate the generalizability of our proposed method, we conducted experiments on the ICDAR-Art dataset, which includes a large number of real-world instances of curved text. As shown in Table \ref{tab7}, our method outperforms current state-of-the-art models. Specifically, the F-measure of our method is 0.81$\%$ higher than  EMA \cite{zhao2022mixed} and 6.71$\%$ higher than the newer method \cite{wang2022fuzzy}.
\subsection{Challenges}\label{subsec5}
While our methodology exhibits robust performance in detecting artistic-style text, it encounters challenges when dealing with more extreme textual variations. As shown in Fig. \ref{FIG_8}(a) and (b), our method still needs to be improved when dealing with text with extreme shapes and highly blended colors with the background. As shown in Fig. \ref{FIG_8}(c), there are few cases of error detection, categorizing non-artistic-style text errors as artistic style titles. Meanwhile, in Fig. \ref{FIG_8}(d), the severe occlusion of the character image resulted in incomplete text detection. We invite you to come up with innovative solutions to meet these challenges.
\section{Discussion}\label{sec5}
Current text detection methods primarily focus on general scenarios, with few addressing the detection of artistic-style text. However, the importance of detecting artistic-style text in our lives has become increasingly apparent. We collected 1,500 movie posters featuring various artistic-style titles to address the current market's lack of artistic-style text data, and we conducted a comprehensive data analysis of this dataset. We evaluated the complexity of each text region based on the number of coordinate points. As shown in Fig. \ref{FIG_6}(a), the complexity is roughly divided into three categories: Complex, Moderately Complex, and Extremely Complex. The complexity distribution in this dataset is relatively balanced. Fig. \ref{FIG_6}(b) shows the area distribution of all annotated text boxes, revealing a wide range of text region sizes in our dataset, including various irregularly shaped artistic-style texts. This further proves the representativeness of our dataset. While our method represents a significant step forward in detecting artistic-style text, Sec. \ref{subsec5} discusses areas for improvement. We hope to achieve more efficient and faster detection of artistic-style text in the future.
\section{Conclusion}\label{sec6}
This paper proposes a novel method that effectively enhances the model's perceptual capabilities to accurately detect artistic-style text with complex structures, filling the gap in current text detection algorithms for detecting artistic-style text. In our method, the RCCA module, which consists of the Criss-Cross Attention Module, effectively utilizes global contextual information to enable the model to capture detailed features that are highly similar to the background. R-FPN based on FPN and residual dense block suppresses the effect of background noise and avoids treating non-text pixels as text pixels. BDM ensures the correctness of boundary modelling. We propose the Movie-Poster dataset to fill the market's gap in artistic-style text data. Extensive experiments demonstrate that our proposed method performs superiorly on the Movie-Poster dataset and produces excellent results on multiple benchmark datasets.
\section*{Acknowledgements}
This work was supported in part by the Chongqing University of Technology high-quality development Action Plan for of graduate education gzlcx20243151, the Basic and Applied Basic Research Foundation of Guangdong Province under Project 2021A1515110298, and in part by the Science and Technology Program of Nansha under Project 2021ZD003

	\bibliographystyle{model5-names}

	\bibliography{cas-refs}




\end{sloppypar}
\end{document}